\documentclass[sigconf]{acmart}

\usepackage{amsmath,amssymb,booktabs,multirow,array,xspace}
\usepackage{xcolor}
\usepackage{pifont}
\usepackage{dblfloatfix}
\usepackage{placeins}
\usepackage{float}
\usepackage{booktabs}
\renewcommand\footnotetextcopyrightpermission[1]{}
\newcommand{\method}{UniMoMo\xspace}
\newcommand{\methoda}{UniMoMo-A\xspace}
\newcommand{\best}[1]{\textbf{#1}}
\newcommand{\cmark}{\ding{51}}

\title{UniMoMo: Expert Merging-Based MoE Acceleration for Large Recommendation Models}
\author{Lei Xin}
\authornote{Equal contribution.}
\email{2835838600@qq.com}
\affiliation{%
  \institution{Kuaishou Technology}
}

\author{Bin Gu}
\authornotemark[1]
\affiliation{%
  \institution{Hohai University}
}

\author{Peize Li}
\authornotemark[1]
\affiliation{%
  \institution{Kuaishou Technology}
}

\author{Zitong Wang}
\affiliation{%
  \institution{Wuhan University}
}

\author{Jianbo Zhao}
\affiliation{%
  \institution{Independent Researcher}
}

\author{Changjiang Jiang}
\affiliation{%
  \institution{Wuhan University}
}

\author{Chao Huang}
\affiliation{%
  \institution{ByteDance, Douyin}
}

\author{Xuyang Zhao}
\affiliation{%
  \institution{Kuaishou Technology}
}

\author{Zunhai Su}
\affiliation{%
  \institution{The University of Hong Kong}
}

\author{Fanhu Zeng}
\authornote{Corresponding author.}
\affiliation{%
  \institution{Independent Researcher}
}

\author{Zhenglun Kong}
\authornotemark[2]
\affiliation{%
  \institution{Northeastern University}
}
\begin{abstract}
Sparse mixture-of-experts (MoE) layers expand recommendation capacity through conditional computation, yet a trained checkpoint still stores and routes over its full expert bank. We study a deployment problem: convert that checkpoint to a smaller standard MoE under an explicit expert budget, without adding a compression-specific online module. 

To address this, we introduce \method, a post-training compression framework formulated as a constrained graph coarsening problem. Rather than relying on parameter distance, \method groups experts based on their functional similarity, using an unlabeled calibration set to measure how similarly experts respond to shared recommendation states. To prevent performance degradation, we introduce a layer-adaptive protection mechanism that restricts the merging of high-traffic experts based on their routing exposure.
Across Amazon Beauty, KuaiRec, and TenRec with 2, 4, and 6 MoE blocks, the final four-expert checkpoints obtain source-relative five-run mean NDCG@10 ratios of 99.92\%--102.30\% and measured A100 speedups of 1.28$\times$--1.63$\times$. An aggressive two-expert, top-1 operating point obtains ratios of 98.36\%--104.24\% and speedups of 1.47$\times$--2.21$\times$. These endpoint results evaluate the complete conversion-and-adaptation workflow and show that a trained recommendation MoE can be exported at multiple serving budgets.
\end{abstract}

\ccsdesc[500]{Information systems~Recommender systems}
\ccsdesc[300]{Computing methodologies~Neural networks}
\ccsdesc[100]{Computing methodologies~Learning latent representations}

\keywords{recommender systems, mixture of experts, model compression, expert merging, adaptive routing}

\begin{document}
\maketitle

\section{Introduction}
Industrial recommenders need model capacity, but serving cost determines which capacity can be deployed. Wukong, DHEN, RankMixer, and TokenMixer-Large expand interaction depth or sparse parameter capacity through accelerator-friendly blocks~\cite{zhang2024wukong,zhang2022dhen,zhu2025rankmixer,jiang2026tokenmixer}. This scaling direction creates a checkpoint-level question that architecture design does not answer: after a sparse recommender has been trained, how much of its expert bank is necessary for serving?
\begin{figure}[t]
  \centering
  \includegraphics[width=\columnwidth]{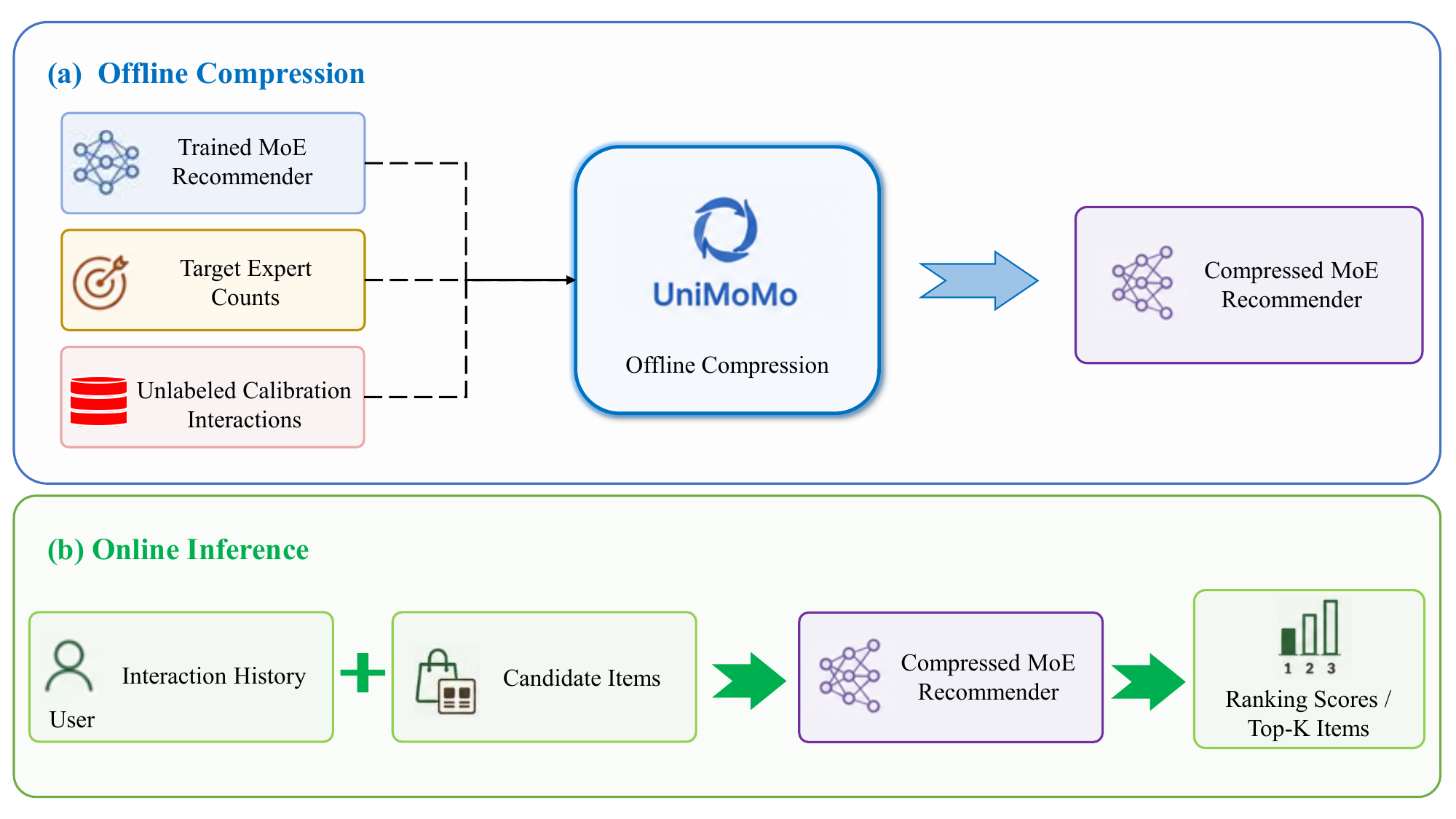}
  \caption{Input–output overview. a) Offline, UniMoMo compresses a trained MoE recommender using unlabeled calibration interactions and a target expert budget. b) Online, the compressed model takes user histories and candidate items to produce ranking scores or top-K recommendations.}
  \Description{A two-stage overview. The upper offline stage takes a trained MoE recommender, target expert counts, and unlabeled calibration interactions as input to UniMoMo and returns a compressed MoE recommender. The lower online stage combines a user's interaction history and candidate items, scores them with the compressed recommender, and returns ranking scores or top-K items.}
  \label{fig:motivation}
\end{figure}

Sparse MoE separates stored capacity from activated computation by routing each token to only a few experts~\cite{jacobs1991adaptive,shazeer2017outrageously,lepikhin2020gshard,fedus2022switch,du2022glam}. Recommendation-specific MoEs use this mechanism for long behavior sequences, heterogeneous interaction frequencies, and multitask learning~\cite{lin2026mixture,zhang2025frequency,zhang2026smes}. These methods design and train new sparse architectures. Systems such as MegaBlocks and ScatterMoE optimize sparse execution but retain the trained expert bank~\cite{gale2023megablocks,tan2024scattered}. Neither line converts an existing recommendation checkpoint to a chosen deployment budget.

Expert reduction is a partitioning decision on the recommendation distribution. Relying on parameter distance is not enough because it does not reveal whether two experts respond similarly to the specific hidden states produced by the recommender. Output similarity alone is also incomplete because the same reconstruction error has a larger downstream footprint when it repeatedly affects a high-traffic expert. A deployment partition must therefore account for both behavioral compatibility and routing exposure under one expert-count constraint.

To bridge this gap, we propose \method, a checkpoint converter that solves this budgeted partition as calibration-conditioned constrained coarsening. 
We formulate this conversion as a graph coarsening process guided by actual expert behavior and routing traffic. 
First, rather than comparing parameter weights, we evaluate all experts on shared calibration data to measure their functional similarity, which forms an expert-affinity graph. 
Second, to prevent performance drops, we use the routing entropy of each layer to identify and protect high-traffic experts from being repeatedly merged. Finally, we reconstruct the new experts by fusing their parameters based on their routing traffic.
Because simple weight averaging is ineffective for non-linear SwiGLU blocks, we apply a least-squares correction to fix any mismatch in the intermediate activations. This offline merge plan uses only unlabeled interactions. After a brief supervised adaptation stage, the compressed model operates with a resized router and standard top-$k$ inference, allowing it to drop directly into existing deployment pipelines.

Experiments on Amazon Beauty, KuaiRec, and TenRec cover 2, 4, and 6 MoE blocks and two deployment budgets. Across the nine paired settings, the four-expert checkpoint obtains 99.92\%--102.30\% of the Origin MoE five-run mean NDCG@10 and runs 1.28$\times$--1.63$\times$ faster. The two-expert operating point obtains 98.36\%--104.24\% and runs 1.47$\times$--2.21$\times$ faster.

Our contributions are:
\begin{itemize}
  \item We define checkpoint conversion for recommendation MoEs as an expert-budgeted deployment problem whose output remains a standard sparse MoE.
  \item We formulate the conversion as one traffic-conditioned coarsening process whose affinities, admissibility rules, and reconstruction all come from the same calibration stream. The output is a smaller standard MoE with no added online module.
  \item Across three datasets and three depths, the default conversion produces source-relative mean NDCG@10 ratios of 99.92\%--102.30\% with 1.28$\times$--1.63$\times$ measured speedup.
\end{itemize}

\section{Related Work}
\subsection{Large recommendation model progress}
Large recommendation models have progressed from factorization and neural matching to cross networks, attention-based interaction layers, and sequence encoders~\cite{koren2009matrix,rendle2012bpr,cheng2016wide,guo2017deepfm,he2017neural,wang2017deep,wang2021dcn,song2019autoint,zhou2018deep,zhou2019deep,hidasi2015session,tang2018personalized,kang2018self,sun2019bert4rec,li2020time,zhou2022filter,de2021transformers4rec}. Recent architectures scale the recommendation backbone itself: Wukong and DHEN deepen feature interaction, while RankMixer and TokenMixer-Large emphasize accelerator-friendly token mixing and sparse per-token capacity~\cite{zhang2024wukong,zhang2022dhen,zhu2025rankmixer,jiang2026tokenmixer}. Together, these models establish the high-capacity ranking workloads for which serving efficiency becomes a first-order constraint.

Sparse experts further expand recommendation capacity by matching computation to heterogeneous user behavior. Multi-gate MoE and progressive layered extraction allocate capacity across recommendation objectives~\cite{ma2018modeling,tang2020progressive}; MoS routes themed behavior subsequences, frequency-aware experts specialize on interaction regimes, and SMES studies expert sparsity for scalable multitask recommendation~\cite{lin2026mixture,zhang2025frequency,zhang2026smes}. This line demonstrates the value of expert specialization inside recommendation models, but determines expert organization during architecture design or training. It leaves a deployment question after training: how can an existing recommendation-MoE checkpoint be resized to a requested serving budget while retaining its ranking utility? Recommendation states, traffic skew, and ranking objectives jointly determine which expert capacity remains valuable at serving time.

\subsection{Recommendation MoE acceleration}
MoE acceleration for large recommendation models spans conditional activation, traffic assignment, and sparse execution. A router activates only a small subset of experts for each state, extending adaptive gating while separating stored capacity from activated computation~\cite{jacobs1991adaptive,shazeer2017outrageously}. For recommendation deployment, the resulting bottlenecks are load imbalance, expert underutilization, dispatch overhead, and device communication; large sparse architectures establish their routing and execution foundations~\cite{lepikhin2020gshard,fedus2022switch,du2022glam,zoph2022st,lewis2021base,rajbhandari2022deepspeed}.

Existing acceleration mechanisms primarily change routing or execution. Balance and specialization objectives, expert-choice, soft, or hash routing, and input-adaptive computation alter how traffic reaches experts~\cite{dai2024deepseekmoe,zhang2022moefication,zoph2022designing,zhou2022mixture,puigcerver2024sparse,liu2024deepseek,raposo2024mixture,roller2021hash}. Gated feed-forward blocks define the expert computation~\cite{shazeer2020glu}; MegaBlocks and ScatterMoE reduce its sparse execution overhead~\cite{gale2023megablocks,tan2024scattered}, while shared experts and different granularities of conditional computation improve capacity utilization~\cite{antoniak2024mixture,jiang2024mixtral}. For recommendation serving, these mechanisms make a chosen sparse architecture more efficient, but retain the expert identities stored in the trained checkpoint. Recommendation MoE acceleration therefore still lacks a checkpoint-level mechanism that removes redundant expert capacity under an explicit serving budget. This axis complements routing and kernel optimization by changing the deployed expert bank itself.


\subsection{Post-training expert-bank compression}
Post-training compression provides the closest checkpoint-level context. Pruning and quantization reduce weights within a fixed network~\cite{frantar2023sparsegpt,sun2024simple,frantar2022gptq,kong2025token}, whereas model soups, task arithmetic, TIES-Merging, and related fusion combine independently trained checkpoints or updates~\cite{wortsman2022model,ilharco2022editing,yadav2023ties,yu2024language,zeng2025robustmerge,guo2025hide}. Neither objective determines how experts trained jointly under one recommendation router should share a smaller bank. In this setting, redundancy depends on expert responses to recommendation states, and deployment risk depends on the traffic routed to each expert.

Expert-bank reduction directly studies jointly routed experts. Merging Experts into One approximates the combined output of activated experts~\cite{he2023merging}; HC-SMoE groups experts through an output-based hierarchy, and MergeMoE combines routing statistics with least-squares reconstruction~\cite{chen2024retraining,miao2025mergemoe}. Other reducers use output subspaces, sparse element-wise fusion, lightweight expert replacement, or cross-layer routing trajectories~\cite{li2026sub,zhao2025puzzlemoe,zhang2025mone,yang2025moe}. These techniques identify useful reduction operations, yet their objectives do not define the recommendation deployment decision addressed here: selecting a target bank size while accounting jointly for behavior on recommendation states and exposure under recommendation traffic. A recommendation-specific criterion must therefore couple functional compatibility with empirical exposure instead of treating every expert pair as equally consequential.

\method formulates a recommendation-native deployment objective as one recommendation-conditioned, expert-budgeted partition defined directly on ranking states and serving traffic. Shared unlabeled interactions define functional edge weights, identify high-exposure experts that should not be modified repeatedly, and weight expert reconstruction. Bregman clustering and graph reduction provide optimization tools~\cite{banerjee2005clustering,von2007tutorial,loukas2019graph}; the contribution is the traffic-conditioned conversion problem and its unified partition. The resulting checkpoint preserves standard top-$k$ routing and adds no compression-specific online module.

\section{Method}
\subsection{Preliminaries}
\paragraph{Sparse recommendation MoE.}
Consider a recommendation model with $L$ sparse MoE blocks. Block $\ell$ contains $E_\ell$ SwiGLU experts $f_{\ell,e}$ and a router. For an input token $x$, the router produces logits $z_\ell(x)$. The active set and its normalized routing weights are
\begin{equation}
 \mathcal{A}_\ell(x)=\operatorname{TopK}(z_\ell(x),k),\qquad
 a_{\ell,e}(x)=\frac{\exp z_{\ell,e}(x)}
 {\sum_{j\in\mathcal{A}_\ell(x)}\exp z_{\ell,j}(x)}.
 \label{eq:routing}
\end{equation}
Only experts in $\mathcal{A}_\ell(x)$ are evaluated during ordinary inference. Their weighted output is
\begin{equation}
  y_\ell(x)=\sum_{e\in\mathcal{A}_\ell(x)}a_{\ell,e}(x)f_{\ell,e}(x).
  \label{eq:moe-output}
\end{equation}

\paragraph{Post-training expert compression.}
Given a trained model and an unlabeled calibration set $\mathcal{C}$, we replace the $E_\ell$ experts with $M_\ell<E_\ell$ experts. A merge plan is a partition $\mathcal{G}_\ell=\{C_{\ell,1},\ldots,C_{\ell,M_\ell}\}$ of the original expert indices. Each cluster becomes one expert in the compressed layer, and the corresponding router rows are reduced from $E_\ell$ to $M_\ell$. Here, \emph{post-training} means that compression starts from a completed backbone checkpoint; it does not mean training-free. Calibration interactions determine the partition and initialize the compressed weights, after which a short, matched supervised fine-tuning stage adapts the complete compressed model.

We seek one partition that is behaviorally coherent while limiting repeated modification of experts with high routing exposure. Parameter distance does not encode either property on the recommendation distribution. A common calibration pass defines both the coarsening objective and its admissible partitions.

\paragraph{Deployment objective.}
For a expert $f_{\ell,e} \in \mathrm{SwiGLU}(d, h)$, the three projections contain approximately $3dh$ parameters. Reducing a layer from $E_\ell$ to $M_\ell$ experts therefore removes $3dh(E_\ell-M_\ell)$ parameters before the comparatively small router update. The end-to-end reduction is
\begin{equation}
 r_P=1-\frac{P_{\mathrm{fixed}}+\sum_{\ell}3dhM_\ell}{P_{\mathrm{fixed}}+\sum_{\ell}3dhE_\ell},
 \label{eq:paramratio}
\end{equation}
where $P_{\mathrm{fixed}}$ includes embeddings, mixing layers, and prediction heads. Eq.~\ref{eq:paramratio} shows why same expert compression ratio can yield different end-to-end parameter reductions across datasets. Activated expert computation depends on the post-compression routing width $k'$, while router selection and dispatch also depend on $M_\ell$.

The target $M_\ell$ is a serving budget, not a hyperparameter selected to maximize a ranking metric. Given that budget, the conversion minimizes disruption to the routed expert functions; source-relative ranking quality is measured after conversion rather than imposed as an algorithmic constraint. We therefore report NDCG, HR, AUC, parameters, and latency separately. This distinction matters when item embeddings dominate total parameters even though a smaller expert bank reduces the work inside every MoE block.

\subsection{Overall framework}
\begin{figure*}[t]
  \centering
  \includegraphics[width=\textwidth]{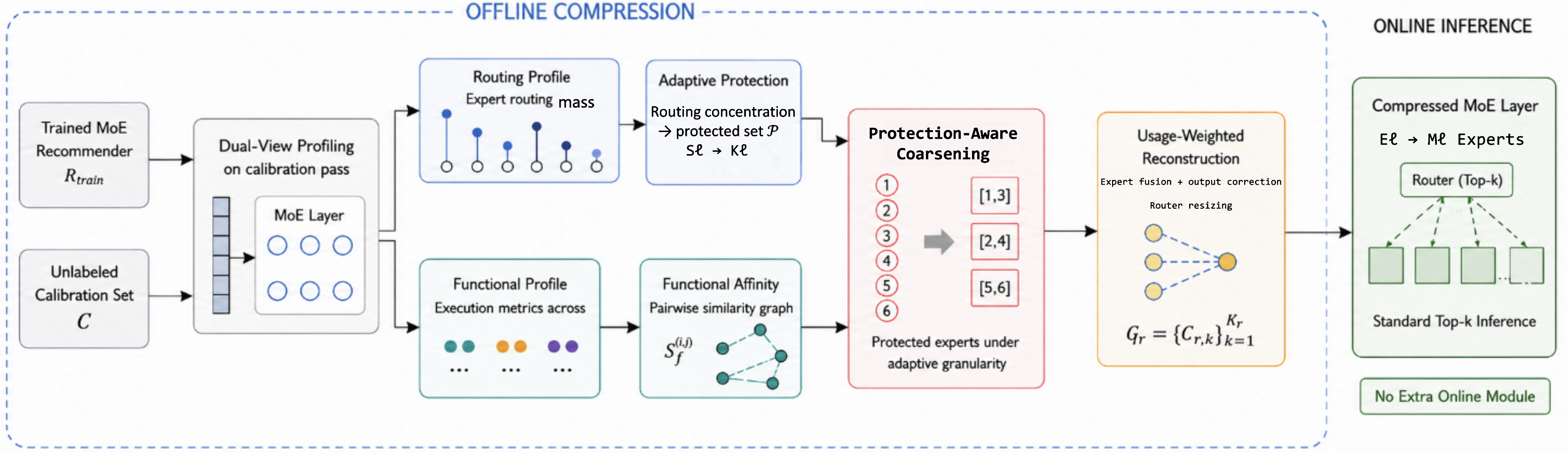}
  \caption{Overview of \method. Given a trained MoE recommender and unlabeled calibration interactions, dual-view profiling records routing traffic and expert-output statistics. The traffic profile sets merge admissibility, while the functional profile defines edge weights in the expert-affinity graph. Graph coarsening forms the groups; traffic-weighted expert reconstruction and arithmetic router-row aggregation produce the compressed layer. The result retains standard top-$k$ routing and introduces no method-specific online module.}
  \Description{The post-training procedure begins with a trained MoE recommender and an unlabeled calibration set. Profiling produces routing-exposure statistics and a functional expert-affinity graph. Exposure controls admissible graph merges. Each resulting group is replaced by one traffic-weighted expert, and the router rows are aggregated arithmetically. The compressed layer uses standard top-k routing and adds no method-specific inference module.}
  \label{fig:framework}
\end{figure*}

Figure~\ref{fig:framework} summarizes the merge-initialization procedure and the inference path of the compressed model. The procedure takes a trained MoE recommender, an unlabeled calibration set $\mathcal{C}$, and a target expert count $M_\ell$ for each layer. It returns the same backbone with a smaller expert bank and a resized router, which is then adapted under the matched supervised protocol in Section~\ref{sec:experiments}.

\paragraph{Calibration graph and exposure constraints.}
For each frozen layer, one pass records all information needed to define the partition. The original top-$k$ decisions provide probability-weighted routing mass $u_{\ell,e}$. On the same valid tokens, every expert is also evaluated to obtain response summaries $\{\mu_{\ell,e},v_{\ell,e}\}$. Shared inputs keep pairwise differences attributable to expert behavior rather than different routed input distributions. The response summaries define affinities $s_{ij}$; the normalized concentration of routing mass determines which high-exposure vertices should not be repeatedly coarsened.

\paragraph{Constrained coarsening.}
\method constructs $\mathcal{G}_\ell=\{C_{\ell,m}\}_{m=1}^{M_\ell}$ by repeatedly combining the most similar admissible pair. Admissibility is part of the same partitioning problem: two protected vertices cannot collapse into one group, and a protected vertex cannot be modified repeatedly. Section~\ref{sec:functional} specifies the greedy solver and the fallback used when the target count conflicts with the initial constraint budget.

\paragraph{Usage-weighted reconstruction.}
Each cluster $C_{\ell,m}$ is replaced by one SwiGLU expert. Routing mass weights the source projections according to their observed contribution to the layer output. A calibration-derived least-squares map then corrects the intermediate activation mismatch caused by averaging nonlinear experts. The resized router uses the arithmetic centroid of the source rows, as defined in Eq.~\ref{eq:router-merge}; routing weights are not reused for router initialization. All corrections are folded into stored parameters.

\paragraph{Compressed inference.}
The compressed layer contains $M_\ell$ experts and uses routing width $k'$. It does not retain the affinity graph, output summaries, entropy calculation, or least-squares operator. Eq.~\ref{eq:moe-output} remains the inference computation, with the router selecting among the compressed experts. Existing top-$k$ routing and expert kernels execute the layer without a \method-specific module.

\paragraph{Compression procedure.}
For each layer, the implementation (1) captures at most 20 batches of valid input states, (2) evaluates every expert on those shared states and accumulates $\mu_{\ell,e}$, $v_{\ell,e}$, and routing mass, (3) computes the protected set, (4) greedily coarsens the affinity graph to $M_\ell$ clusters, and (5) reconstructs the experts and router. After all layers are replaced, the complete compressed recommender is fine-tuned and selected by validation NDCG@10. Test interactions are excluded from every step.

\subsection{Recommendation-aware coarsening and reconstruction}
\label{sec:functional}
Given calibrated expert responses and routing exposure, \method first solves for a constrained partition and then reconstructs one expert for each cluster.

\paragraph{Functional profiling.}
Parameter proximity does not guarantee similar responses on recommendation interactions. Two experts can move to distant parameter regions during training yet remain close on hidden states produced by the recommender. Conversely, small parameter distance does not prevent large output shift after nonlinear gating. We compare responses that enter the residual stream.
To ensure a fair comparison of expert behaviors, we bypass the router and evaluate all experts on the same set of $N$ valid calibration tokens, $\{x_t\}_{t=1}^{N}$, at the input of layer $\ell$. Let $o_{\ell,e,t}=f_{\ell,e}(x_t)$ be the output of expert $e$. To avoid the memory overhead of storing all outputs, we summarize each expert's behavior as an isotropic Gaussian distribution, parameterized by a mean vector $\mu_{\ell,e}$ and a variance scalar $v_{\ell,e}$:
\begin{equation}
 \mu_{\ell,e}=\frac{1}{N}\sum_{t=1}^{N}o_{\ell,e,t},\qquad
 v_{\ell,e}=\frac{1}{dN}\sum_{t=1}^{N}\lVert o_{\ell,e,t}-\mu_{\ell,e}\rVert_2^2.
\end{equation}
We then measure the functional discrepancy between expert $i$ and expert $j$ using the KL divergence between their Gaussian summaries, denoted as $q(i\Vert j)$:
\begin{equation}
 q(i\Vert j) = \frac{d}{2}\left( \frac{v_i}{v_j} - 1 + \log\frac{v_j}{v_i} \right)+ \frac{\lVert\mu_i-\mu_j\rVert_2^2}{2v_j}.
\end{equation}
This formulation intuitively separates the discrepancy into the difference in output scales and the Euclidean distance between their mean behaviors. Since KL divergence is asymmetric, we define a symmetric distance $D_{ij}$ and map it to an affinity score $s_{ij} \in (0, 1]$:
\begin{equation}
 D_{ij} = \frac{1}{2}\big(q(i\Vert j) + q(j\Vert i)\big),\qquad
 s_{ij}=\frac{1}{1+D_{ij}}.
\end{equation}
Identical experts yield $s_{ij}=1$, with the score decreasing monotonically as their behaviors diverge. To prevent numerical instability, a small constant $\epsilon$ is added to $v_{\ell,e}$ before computation. The complete graph for layer $\ell$ is thus constructed with one node per expert and edge weights $s_{ij}$.

Shared inputs are necessary for this comparison. If expert $i$ and expert $j$ were summarized only on their routed tokens, $D_{ij}$ would mix expert behavior with different input distributions. Running every expert on $\{x_t\}$ holds that distribution fixed. The isotropic summary is practical for wide recommendation layers: a full covariance would require $O(E_\ell d^2)$ storage, whereas the mean and scalar variance require $O(E_\ell d)$ and can be accumulated batch by batch.

The two statistics serve different purposes. The squared mean difference penalizes a persistent shift in the expert output, while the variance terms distinguish experts with different response scales. Since the graph is used to order merge candidates, $s_{ij}$ does not need a probabilistic interpretation or a globally tuned threshold.

\paragraph{Constrained graph coarsening.}
We initialize each expert as a singleton cluster. For clusters $A$ and $B$, average-link similarity is
\begin{equation}
 \operatorname{sim}(A,B)=\frac{1}{|A||B|}\sum_{i\in A}\sum_{j\in B}s_{ij}.
\end{equation}
At each step, the algorithm merges the valid pair with the highest similarity until $M_\ell$ clusters remain. The protected set $\mathcal{P}_\ell$, defined in Section~\ref{sec:adaptive}, controls validity. Two protected clusters cannot merge. A protected singleton may absorb one unprotected cluster, but the resulting cluster is removed from further consideration. The restriction lets a high-traffic expert represent one nearby cluster without repeated modification.

The protected count can exceed the target count in a strongly concentrated layer. If no constrained pair remains while more than $M_\ell$ clusters are active, the implementation disables protection for that merge only and selects the highest-similarity pair under the same average-link score. It then restores the constraint for the next step. Protection is therefore a deterministic merge priority rather than a hard constraint that can make the target infeasible.

Average-link similarity reflects the aggregate compatibility of two groups. It avoids the chaining behavior of single link and the sensitivity of complete link to one atypical pair. Ties are resolved by expert index, making the merge plan deterministic for a fixed checkpoint and calibration stream. We refer to this procedure as greedy graph coarsening because the implementation does not compute a graph Laplacian or preserve its spectrum.

\paragraph{Usage-weighted reconstruction.}
The partition specifies membership, but we must still determine the weights of the newly merged experts. A uniform parameter average would give a rarely selected expert the same influence as an expert that supplies a large fraction of the layer output. Therefore, for each cluster $C_m$, we first compute a traffic-weighted average of the parameters based on the empirical routing mass $u_e$:
\begin{equation}
 w_e=\frac{u_e}{\sum_{j\in C_m}u_j},\qquad
 \bar W_m^r=\sum_{e\in C_m}w_e W_e^r,\quad r\in\{U,G,D\},
 \label{eq:weighted-merge}
\end{equation}
where $U$, $G$, and $D$ denote the up, gate, and down projections. 
$\bar W_m^r$ provides the initial SwiGLU parameters. 

However, a direct parameter fusion suffers from the non-linearity of the SwiGLU block: parameter averaging and non-linear activation do not commute.
Let $X\in\mathbb{R}^{N\times d}$ store the calibration states as rows. If we average the intermediate activations of the original experts, we obtain a better target activation $Q\in\mathbb{R}^{N\times h}$:
\begin{equation}
 Q=\sum_{e\in C_m}w_e\left[\operatorname{SiLU}(X(W^G_e)^\top)\odot X(W^U_e)^\top\right].
\end{equation}
However, the actual intermediate activation (denote $P\in\mathbb{R}^{N\times h}$) produced by $\bar W_m^U$ and $\bar W_m^G$ will generally differ from $Q$.  To correct this mismatch, we find a linear transformation $T_m$ that maps $P$ to $Q$ via least-squares:
\begin{equation}
 T_m=\arg\min_{T\in\mathbb{R}^{h\times h}}\lVert PT-Q\rVert_F^2=P^{\dagger}Q
 \label{eq:activation-correction}
\end{equation}
The Moore--Penrose pseudoinverse $P^{\dagger}$ ensures a minimum-norm solution even if $P$ is rank-deficient. We then fold this correction directly into the merged down projection by setting $W_m^D\leftarrow\bar W_m^D T_m^\top$. This correction is absorbed into the stored weights and adds no operator overhead to the forward pass.

Finally, let $r_{\ell,e}$ denote row $e$ of the original router matrix. The router row for cluster $C_m$ is initialized as
\begin{equation}
 \bar r_{\ell,m}=\frac{1}{|C_m|}\sum_{e\in C_m}r_{\ell,e}.
 \label{eq:router-merge}
\end{equation}
This arithmetic centroid is a neutral initialization: unlike a traffic-weighted row, it does not let the calibration-frequency estimate enter both grouping and router initialization. It is not expected to preserve the original cluster log-sum-exp exactly. The resized router produces one logit per merged cluster, so no mapping from old expert indices is needed at inference. Supervised adaptation updates the compressed router, merged experts, and the remaining model parameters; the calibration correction supplies its starting point rather than replacing adaptation.

\subsection{Routing-exposure merge priority}
\label{sec:adaptive}
Behavioral similarity finds experts that can be merged, but it ignores the risk of modifying frequently used experts. If an expert processes a large fraction of tokens, any error in its reconstruction will easily spread to later layers. Therefore, routing frequency must help decide which experts are safe to merge. Layers where routing traffic is highly concentrated need to protect more experts from being merged.

For a calibration token $x_t$, the router gives a routing weight $a_{\ell,e}(x_t)$ to expert $e$ only if it is selected ($e\in\mathcal{A}_\ell(x_t)$). To preserve the original router's confidence, we calculate the total routing mass $u_{\ell,e}$ and its probability distribution $p_{\ell,e}$ using these soft weights instead of hard counts:
\begin{equation}
 u_{\ell,e}=\sum_{t=1}^{N}\mathbb{I}[e\in\mathcal{A}_\ell(x_t)]a_{\ell,e}(x_t),\qquad
 p_{\ell,e}=\frac{u_{\ell,e}}{\sum_{j=1}^{E_\ell}u_{\ell,j}}.
 \label{eq:routing-mass}
\end{equation}
To measure how unbalanced the routing is within a layer, we compute the normalized entropy deficit $S_\ell$:
\begin{equation}
 S_\ell=1-\frac{-\sum_{e=1}^{E_\ell}p_{\ell,e}\log(p_{\ell,e}+\epsilon)}{\log E_\ell},
 \label{eq:skew}
\end{equation}
where $\epsilon$ is a small number for numerical stability. Dividing by $\log E_\ell$ makes this score independent of the total number of experts, keeping $S_\ell$ strictly between $0$ and $1$. 

We use this imbalance score to determine the fraction of protected experts ($\gamma_\ell$) and the exact protected count ($K_\ell$) for each layer:
\begin{equation}
 \gamma_\ell=\gamma_{\min}+(\gamma_{\max}-\gamma_{\min})S_\ell^{\beta},\qquad
 K_\ell=\left\lceil\gamma_\ell E_\ell\right\rceil.
 \label{eq:adaptive}
\end{equation}

The $K_\ell$ experts with the largest $u_{\ell,e}$ form $\mathcal{P}_\ell$. We use $\gamma_{\min}=0.1$, $\gamma_{\max}=0.4$, and $\beta=1$ in all reported experiments. These values are shared across layers and datasets. The optional sigmoid mapping in the implementation is disabled.

\paragraph{Practical feasibility and limitations.}

For models with $E_\ell=8$ experts, Eq.~\ref{eq:adaptive} guarantees $1\le K_\ell\le4$ protected experts. The standard target $M_\ell=4$, which supports the main claim, is feasible without protection relaxation. The two-expert setting is a separate aggressive operating point; when $K_\ell>2$, it uses the stepwise fallback in Section~\ref{sec:functional}. We do not use that operating point as evidence for a hard-protection claim.

Normalization by $\log E_\ell$ removes the direct dependence of entropy on the number of experts. Under uniform routing, the entropy is $\log E_\ell$, so $S_\ell=0$ and $\gamma_\ell=\gamma_{\min}$. When one expert receives nearly all routing mass, $S_\ell$ approaches one and $\gamma_\ell$ approaches $\gamma_{\max}$. The exponent $\beta$ changes the response between these endpoints. Values above one postpone additional protection until routing becomes strongly concentrated; values below one increase protection earlier. We use the linear map because the current experiments do not tune a transition point.

Eq.~\ref{eq:adaptive} turns the observed routing distribution into a deterministic admissibility schedule inside the coarsening procedure. A balanced layer begins with fewer protected representatives than a concentrated layer with the same $E_\ell$. Behavioral affinity supplies the merge score, while routing exposure determines which candidate pairs remain admissible as the partition is coarsened.

Routing mass does not capture every form of importance. A low-traffic expert can encode a rare but useful behavior, and routing mass alone cannot identify that case. \method reduces the risk by merging according to calibrated behavior rather than traffic, but the calibration set still determines which rare responses are observed. Section~\ref{sec:limitations} discusses this dependence.

\subsection{Reconstruction approximation}
\label{sec:theory}
Our least-squares correction actively fixes the intermediate activation mismatch, rather than directly optimizing the final expert output. To understand the reliability of this design, we can bound the final output error. Let
$H_e=\operatorname{SiLU}(X(W_e^G)^\top)\odot X(W_e^U)^\top$,
so $Q=\sum_e w_eH_e$. The traffic-weighted output of the source cluster and the reconstructed output are
\begin{equation}
 Y_m^\star=\sum_{e\in C_m}w_eH_e(W_e^D)^\top,\qquad
 \widehat Y_m=PT_m(\bar W_m^D)^\top.
\end{equation}
By adding and subtracting the term $Q(\bar W_m^D)^\top$ and applying the triangle inequality, we can bound the difference between these two final outputs:
\begin{align}
 \|Y_m^\star-\widehat Y_m\|_F
 &\leq \|Q-PT_m\|_F\|\bar W_m^D\|_2 \nonumber\\
 &\quad+\sum_{e\in C_m}w_e\|H_e\|_F
       \|W_e^D-\bar W_m^D\|_2.
 \label{eq:reconstruction-bound}
\end{align}
This bound splits the approximation error into two intuitive parts. 
The first term is exactly the residual minimized by Eq.~\ref{eq:activation-correction}, scaled by the merged-down projection. The second term measures dispersion among the source down projections. The initialization is exact when the fitted activation residual vanishes and the source down projections agree; otherwise Eq.~\ref{eq:reconstruction-bound} states the two approximation errors explicitly. The Moore--Penrose solution is most effective when the columns of $P$ span the traffic-weighted activation $Q$. If $P$ is rank deficient, the minimum-norm map leaves the component of $Q$ outside that span in the first residual term. \method preserves the traffic-weighted down-projection initialization and lets supervised adaptation correct the remaining mismatch.

\subsection{Computational complexity}
\paragraph{One-time compression cost.}
For one layer, profiling all $E_\ell$ experts on $N$ calibration tokens costs $O(NE_\ell dh)$ time. The retained moments occupy $O(E_\ell d)$ memory. Constructing the affinity graph costs $O(E_\ell^2d)$ time and $O(E_\ell^2)$ memory. A direct average-link implementation has an $O(E_\ell^3)$ time upper bound when all cluster-pair scores are recomputed after each merge. Solving $PT\approx Q$ by a dense pseudoinverse costs $O(Nh^2+h^3)$ time and $O(Nh+h^2)$ working memory per output cluster. These costs occur offline; activations are released after reconstruction.

\paragraph{Inference-time comparison.}
For one token, an uncompressed layer requires $O(dE_\ell+kdh)$ arithmetic for router scoring and the $k$ activated experts. Its expert and router parameters occupy $O(E_\ell dh+dE_\ell)$ space. After compression, these terms become
\begin{equation}
  O(dM_\ell+k'dh)
  \quad\text{and}\quad
  O(M_\ell dh+dM_\ell),
  \label{eq:complexity}
\end{equation}
respectively, where $M_\ell<E_\ell$. The affinity graph, entropy statistic, and least-squares map are absent from Eq.~\ref{eq:complexity} because they are folded into the stored parameters before evaluation.

The standard configuration keeps $k'=k=2$. Its activated-expert arithmetic therefore has the same asymptotic order as the original layer, while router scoring and parameter storage decrease with $M_\ell/E_\ell$. The smaller expert bank also reduces the number of routing destinations handled by the implementation. \methoda sets $k'=1$ and reduces both router cost and activated-expert arithmetic. Wall-clock latency still depends on the kernel, batch shape, and fixed embedding cost, so we report measured latency rather than infer speedup from complexity alone.

Calibration labels are not used for profiling, routing counts, or reconstruction. Labels enter only during fine-tuning and validation, and test interactions are excluded from every compression decision.

\section{Experiments}
\label{sec:experiments}
The evaluation asks one deployment question: at a fixed source checkpoint, how much expert-bank cost can be removed while retaining its sampled ranking quality? We test this question across datasets, model depths, and two expert budgets. The component study then checks whether both inputs to the partition are useful.

\subsection{Setup}
\textbf{Datasets.} We evaluate Amazon Beauty, KuaiRec, and TenRec~\cite{gao2022kuairec,yuan2022tenrec}. These sources cover markedly different data scales. We use the KuaiRec big matrix rather than mixing its dense matrix into training, and process the selected TenRec interaction file independently. For every source, we retain users with at least five interactions before applying the common temporal protocol below. 

Amazon Beauty contains 2,023,070 interactions before filtering; this yields 52,374 users, 121,291 items, 312,649 training prefixes, and one validation and one test target per user. The KuaiRec big matrix contains 7,176 users, 10,728 items, and 12,530,806 interactions. The complete TenRec collection contains about five million users and 140 million interactions across four recommendation scenarios.

\textbf{Protocol.} User histories are sorted by timestamp. The last two interactions are held out for validation and test, and every earlier prefix produces a training example. Training samples one unseen negative. Sampled evaluation ranks the positive item against 99 negatives and reports HR@10, NDCG@10, and AUC. Within each run, Python, NumPy, and PyTorch share one seed; validation and test candidate lists use deterministic offsets from that seed. All methods reuse the same split and candidate lists. We repeat each experiment five times and report the mean across runs.

For a ranked list $\pi_u$, HR@10 is one when the held-out item occurs in its first ten positions and zero otherwise. NDCG@10 discounts a hit at rank $r$ by $1/\log_2(r+1)$, so it distinguishes early from late hits. AUC measures the fraction of sampled negatives scored below the positive item. We use NDCG@10 for early stopping because it reflects both retrieval and top-rank position.

\textbf{Model and optimization.} The RankMixer backbone uses sequence length 16, hidden size 128, feed-forward multiplier 4, dropout 0.1, eight experts per MoE block, and top-2 routing. The standard target is $E_\ell=8\rightarrow M_\ell=4$ with top-2 routing; the aggressive point uses $8\rightarrow2$ and top-1 routing. With $d=128$, $h=512$, and six blocks, these targets remove 4.72M and 7.08M expert parameters, matching 20.25M and 17.89M totals in Table~\ref{tab:main_results_l6}. We evaluate 2, 4, and 6 blocks; Appendix~\ref{app:backbone-generalization} adds four backbones. AdamW uses learning rate $10^{-3}$, weight decay $10^{-6}$, batch size 512, and at most 20 epochs, with patience three on validation NDCG@10. Compression uses 20 unlabeled training batches. Every baseline receives the same full-model adaptation budget: AdamW at $5\times10^{-4}$ for at most five epochs, with identical training examples and checkpoint selection.

The objective is pointwise binary cross-entropy over one positive and one sampled negative per prefix; an internal variable named ``BPR'' does not denote pairwise BPR loss. Router probabilities weight the selected experts. Apart from the stated target expert count and routing width, backbone, data, optimization, and evaluation settings are unchanged.

\textbf{Baselines.} Origin MoE is the source eight-expert checkpoint. MergeMoE is the matched $8\rightarrow4$ compression baseline and receives the same adaptation budget as \method~\cite{miao2025mergemoe}. Table~\ref{tab:main_results_l6} restricts the numerical comparison to this checkpoint family. ScatterMoE and MoMoE are execution systems, while MoS is a separately trained recommendation architecture~\cite{tan2024scattered,lin2026mixture}; they remain related context rather than checkpoint-conversion baselines. Appendix~\ref{app:backbone-generalization} tests portability across four recommendation backbones.







\begin{table*}[t]
\caption{
Main results with six MoE blocks on Amazon Beauty, KuaiRec, and
TenRec. Higher values are better for ranking metrics, while lower
values are better for parameters and latency. The best and
second-best results within each dataset are shown in bold and
underlined, respectively.
}
\label{tab:main_results_l6}
\small
\renewcommand{\arraystretch}{1.05}

\begin{tabular*}{0.7\textwidth}{
@{\extracolsep{\fill}}
lccccc
}
\toprule
\textbf{Method}
& \textbf{NDCG@10 $\uparrow$}
& \textbf{HR@10 $\uparrow$}
& \textbf{AUC $\uparrow$}
& \textbf{Params (M) $\downarrow$}
& \textbf{Latency (ms)} $\downarrow$ \\
\midrule

\multicolumn{6}{l}{\textit{\textbf{Amazon Beauty}}} \\[-1pt]

ScatterMoE
& 0.2644 & 0.4376 & \underline{0.6795}
& 24.97 & 8.355 \\

MoMoE
& 0.2626 & 0.4372 & 0.6788
& 24.97 & 6.705 \\

MoS
& 0.1081 & 0.2204 & 0.5844
& 29.30 & 4.624 \\

Origin MoE
& 0.2649 & 0.4391 & 0.6778
& 24.97 & 4.833 \\

MergeMoE
& \underline{0.2682}
& \underline{0.4428}
& 0.6781
& \underline{20.25}
& \underline{3.108} \\

\textbf{UniMoMo}
& \textbf{0.2686}
& \textbf{0.4445}
& 0.6785
& \underline{20.25}
& 3.127 \\

\textbf{UniMoMo-A}
& 0.2671
& 0.4420
& \textbf{0.6804}
& \textbf{17.89}
& \textbf{2.345} \\

\midrule
\multicolumn{6}{l}{\textit{\textbf{KuaiRec}}} \\[-1pt]

ScatterMoE
& 0.6962 & \underline{0.9674} & \textbf{0.9798}
& 10.82 & 7.664 \\

MoMoE
& 0.6925 & 0.9635 & 0.9792
& 10.82 & 7.618 \\

MoS
& 0.5610 & 0.8842 & 0.9586
& 15.15 & 5.780 \\

Origin MoE
& 0.6893 & 0.9659 & 0.9793
& 10.82 & 6.860 \\

MergeMoE
& 0.6935
& 0.9657
& 0.9773
& \underline{6.10}
& 4.250 \\

\textbf{UniMoMo}
& \underline{0.6963}
& \textbf{0.9677}
& 0.9796
& \underline{6.10}
& \underline{4.210} \\

\textbf{UniMoMo-A}
& \textbf{0.6964}
& \underline{0.9674}
& \underline{0.9797}
& \textbf{3.74}
& \textbf{3.099} \\

\midrule
\multicolumn{6}{l}{\textit{\textbf{TenRec}}} \\[-1pt]

ScatterMoE
& 0.5932
& 0.7696
& 0.8966
& 162.54
& 10.270 \\

MoMoE
& 0.5941
& 0.7651
& 0.8948
& 162.54
& 9.527 \\

MoS
& 0.5121
& 0.7103
& 0.8834
& 166.87
& \textbf{5.119} \\

Origin MoE
& 0.5938
& 0.7692
& 0.8966
& 162.54
& 9.566 \\

MergeMoE
& 0.5936
& 0.7692
& 0.8961
& \underline{157.82}
& 8.668 \\

\textbf{UniMoMo}
& \underline{0.5987}
& \underline{0.7802}
& \underline{0.8985}
& \underline{157.82}
& 7.451 \\

\textbf{UniMoMo-A}
& \textbf{0.6054}
& \textbf{0.7894}
& \textbf{0.9042}
& \textbf{155.46}
& \underline{6.515} \\

\bottomrule
\end{tabular*}
\end{table*}

\textbf{Efficiency.} We measure inference latency on one NVIDIA A100 80\,GB GPU using BF16 and batch size 256. Each measurement contains 20 warm-up iterations and 80 synchronized timed forward passes. Wall-clock claims compare each \method deployment checkpoint with Origin MoE on the same dataset and depth. The conversion rule is evaluated through ranking quality under the matched adaptation protocol; speed is a property of the resulting $(M_\ell,k')$ serving shape. Total parameter counts include item embeddings, so expert reduction is less visible in the TenRec parameter total than in its MoE execution path.

\begin{table}[t]
\centering
\caption{Matched ablation sweep on Amazon Beauty. Every row compresses $8\!\rightarrow\!4$ with top-2 routing; values are five-run means from this sweep.}
\label{tab:ablation}
\scriptsize
\renewcommand{\arraystretch}{1.05}
\setlength{\tabcolsep}{2.2pt}
\begin{tabular*}{\columnwidth}{@{\extracolsep{\fill}}lccrrr@{}}
\toprule
\textbf{Variant} & \textbf{Beh.} & \textbf{Exp.} & \textbf{NDCG} $\uparrow$ & \textbf{HR} $\uparrow$ & \textbf{AUC} $\uparrow$\\
\midrule
W/o both signals &  &  & 0.2662 & 0.4401 & \best{0.6811}\\
w/o behavior affinity &  & \cmark & 0.2670 & 0.4425 & 0.6801\\
w/o exposure constraint & \cmark &  & 0.2668 & 0.4418 & 0.6794\\
\method & \cmark & \cmark & \best{0.2686} & \best{0.4445} & 0.6785\\
\bottomrule
\end{tabular*}
\end{table}

\subsection{Comparison results}
Table~\ref{tab:main_results_l6} reports the matched six-block checkpoint family. MergeMoE and \method have the same four-expert, top-2 serving shape and the same adaptation budget, so this comparison isolates the quality of the conversion pipeline from the structural source of latency reduction.

\textbf{Matched conversion quality.} The NDCG@10 gains of \method over MergeMoE are 0.0004, 0.0028, and 0.0051 on Amazon Beauty, KuaiRec, and TenRec. We read these values as matched five-run gains rather than a statistical superiority claim. The result needed for deployment is that the traffic-conditioned partition consistently improves upon the matched output-aware compressor while producing a standard four-expert checkpoint.

\textbf{Source-relative endpoints.} Across the nine dataset-depth pairs, the four-expert checkpoints obtain 99.92\%--102.30\% of Origin MoE mean NDCG@10; the two-expert range is 98.36\%--104.24\%. These endpoints include the fixed supervised adaptation stage. Ratios above 100\% are therefore not attributed to compression itself.

\textbf{Serving cost.} The four-expert conversion yields \textbf{1.28$\times$ to 1.63$\times$} speedup. The two-expert, top-1 point yields \textbf{1.47$\times$ to 2.21$\times$}. On TenRec, item embeddings dominate total parameters; latency still changes because every MoE block uses a smaller expert bank.

\subsection{Ablation study}
Table~\ref{tab:ablation} holds the expert bank, routing width, reconstruction, data, and adaptation budget fixed at $8\rightarrow4$ and top-2. ``W/o both signals'' removes calibrated behavior scores and exposure admissibility from the grouping decision while retaining the common conversion scaffold. The next two rows restore one signal at a time. The complete partition has the highest NDCG@10 and HR@10 means in this sweep; relative to the signal-free row, the changes are 0.0018 and 0.0034. This experiment checks whether both inputs are useful in the complete pipeline. It does not use the mean differences as a statistical superiority claim.

\begin{figure}[t]
  \centering
  \begin{minipage}[t]{0.485\columnwidth}
    \centering
    \includegraphics[width=\linewidth,trim=0 0 0 32,clip]{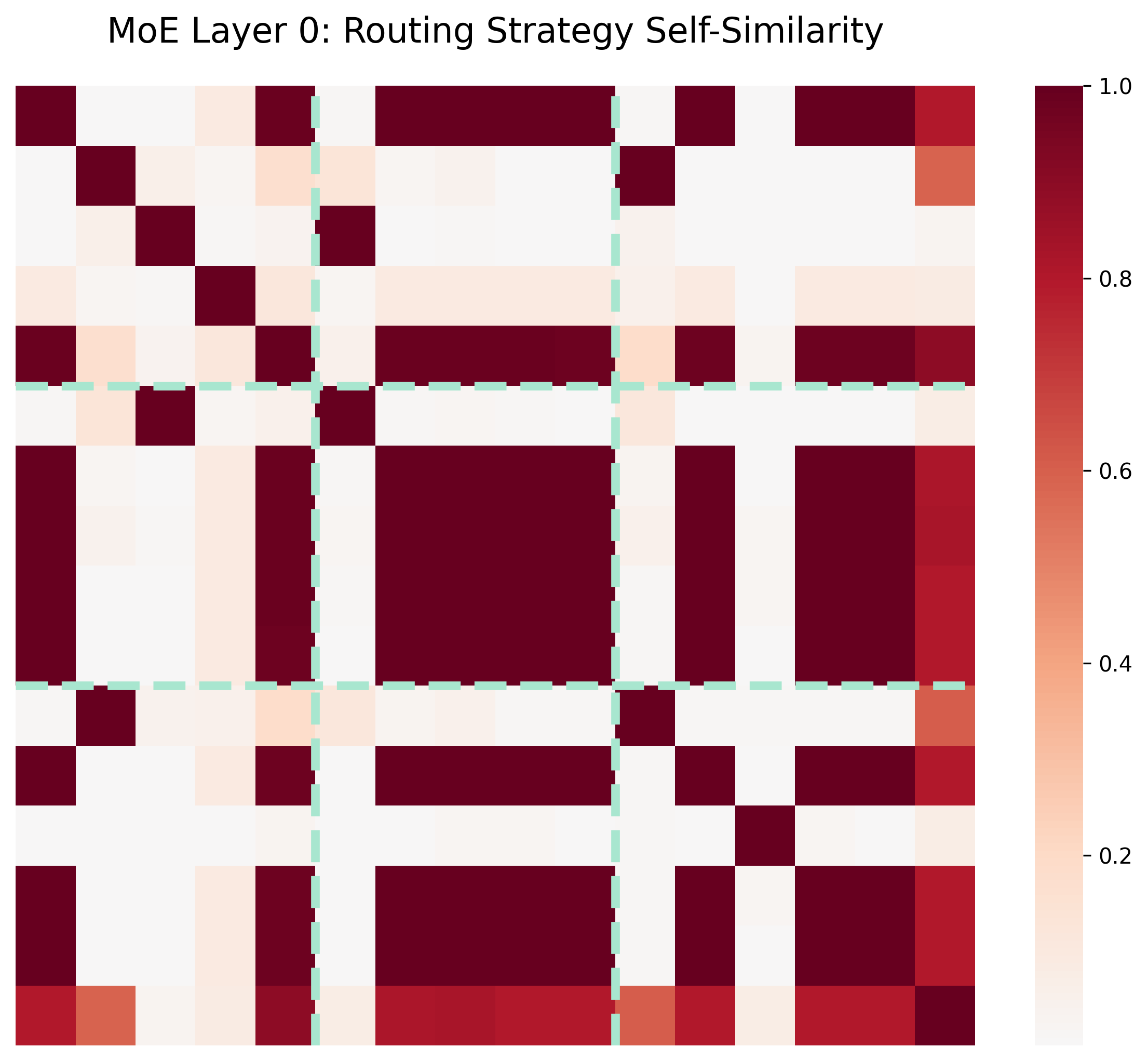}\\[-1mm]
    {\scriptsize (a) ScatterMoE}
  \end{minipage}\hfill
  \begin{minipage}[t]{0.485\columnwidth}
    \centering
    \includegraphics[width=\linewidth,trim=0 0 0 32,clip]{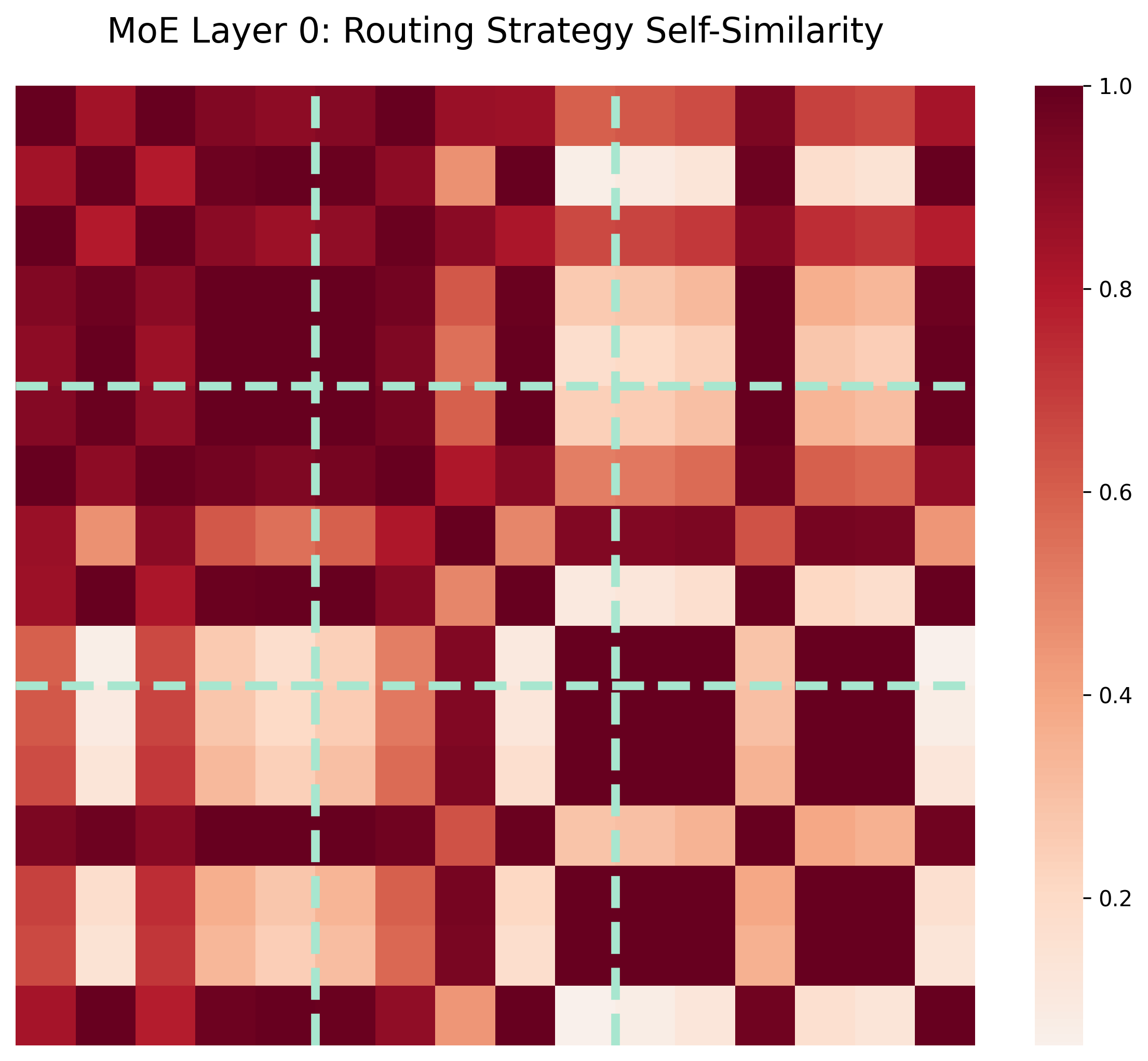}\\[-1mm]
    {\scriptsize (b) \method}
  \end{minipage}
  \caption{Layer-0 routing-pattern diagnostic on the same calibration scale. Each cell compares two routing vectors; darker cells indicate larger similarity, and dashed lines show the displayed routing groups. This diagnostic concerns routing exposure, while Eq.~(4)--(6) define the functional affinity used for coarsening.}
  \Description{Two side-by-side routing-similarity heatmaps. The left panel shows ScatterMoE and the right panel shows UniMoMo. Dark red cells indicate larger similarity, and mint dashed lines divide the displayed routing groups.}
  \label{fig:routing-similarity}
\end{figure}

\subsection{Parameter Sensitivity Analysis}
\begin{figure}[t]
  \centering
  \includegraphics[width=\columnwidth]{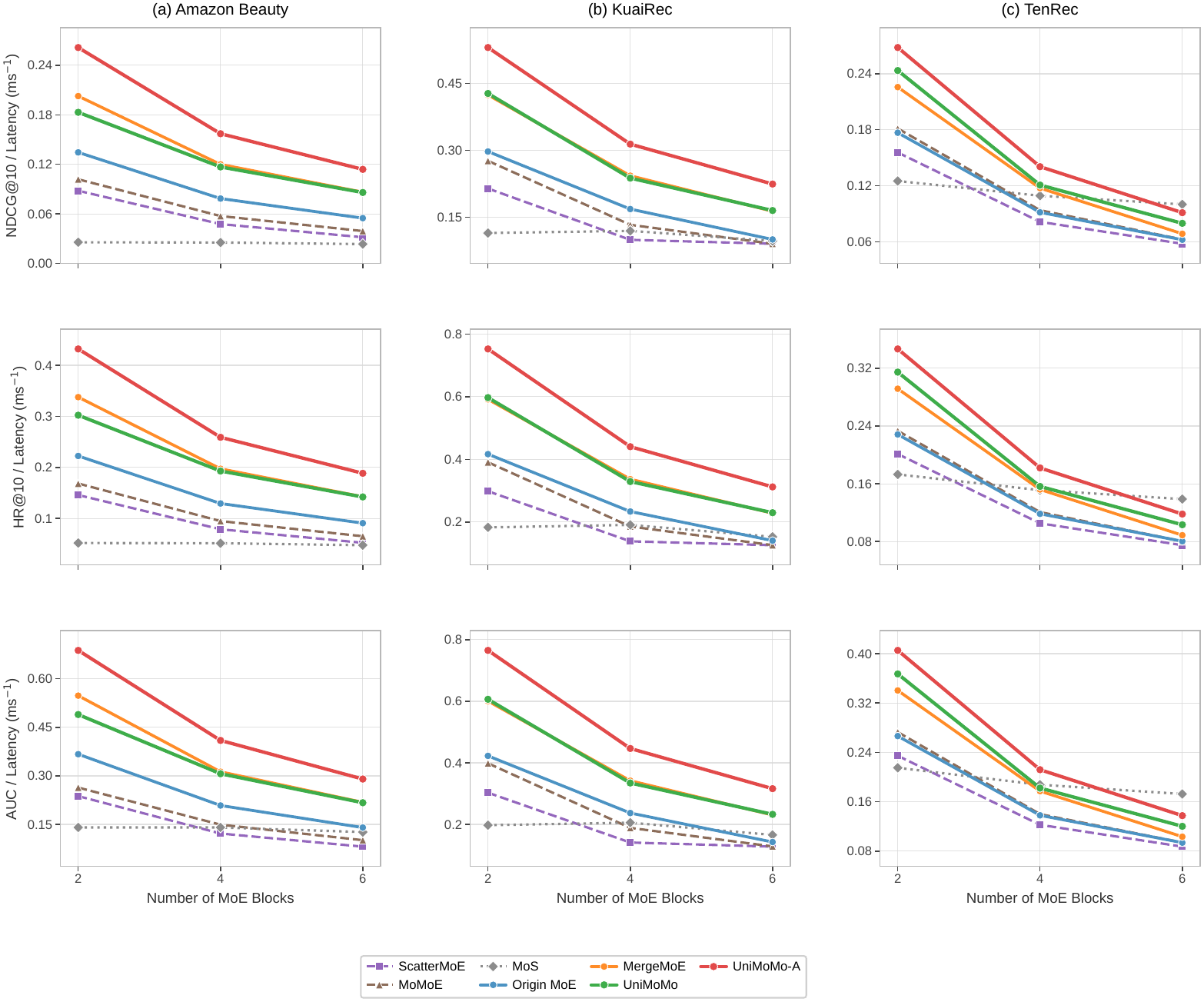}
  \vspace{-0.5cm}
  \caption{Efficiency-normalized performance across MoE block depths. Columns show the three datasets; rows show NDCG@10, HR@10, and AUC divided by latency.}
  \Description{Nine line charts arranged in three columns and three rows. Columns show Amazon Beauty, KuaiRec, and TenRec, and rows show NDCG, HR, and AUC divided by measured latency across different numbers of MoE blocks.}
  \label{fig:parameter-sensitivity}
\end{figure}

\textbf{Comparative Experiments with Different Numbers of Experts.} Four experts is the default setting used in all main-table comparisons. We further evaluate \method with 2, 4, and 8 experts in Table~\ref{tab:expert-sensitivity}. Compared with eight experts, the four-expert default reduces latency by \textbf{40.36\%} while retaining \textbf{99.37\%} of NDCG@10, yielding a \textbf{66.61\%} improvement in NDCG@10 per unit latency. Reducing the count from four to two gives a further 15.77\% latency reduction but lowers NDCG@10 by 1.68\%, confirming four experts as the more balanced default.
\begin{table}[t]
\centering
\caption{Sensitivity to expert count on Amazon Beauty. Four experts is the default; the best value in each metric is bold.}
\label{tab:expert-sensitivity}
\scriptsize
\setlength{\tabcolsep}{3.2pt}
\renewcommand{\arraystretch}{1.02}
\begin{tabular*}{0.95\columnwidth}{@{\extracolsep{\fill}}lrrrr@{}}
\toprule
\textbf{Experts} &
\textbf{NDCG@10} $\uparrow$ &
\textbf{HR@10} $\uparrow$ &
\textbf{AUC} $\uparrow$ &
\textbf{Latency (ms)} $\downarrow$ \\
\midrule
2 & 0.2641 & 0.4375 & 0.6734 & \best{2.634} \\
\textbf{4 (default)} & 0.2686 & 0.4445 & 0.6785 & 3.127 \\
8 & \best{0.2703} & \best{0.4492} & \best{0.6815} & 5.243 \\
\bottomrule
\end{tabular*}
\end{table}

\textbf{Comparative Experiments with Different Numbers of Blocks.} Considering that industrial scenarios require larger parameter scales, we evaluate different MoE depths in Table~\ref{tab:block-sensitivity} and compare their efficiency with existing expert models in Figure~\ref{fig:parameter-sensitivity}. Increasing the depth from two to six blocks improves NDCG@10 by \textbf{6.04\%} and HR@10 by \textbf{6.29\%}. Moving from six to eight blocks yields only a further 0.63\% NDCG@10 gain while increasing latency by 145.70\%, supporting six blocks as the main quality--efficiency operating point.
\begin{table}[t]
\centering
\caption{Sensitivity to MoE depth on Amazon Beauty. Six blocks is the main-table setting.}
\label{tab:block-sensitivity}
\scriptsize
\setlength{\tabcolsep}{3.2pt}
\renewcommand{\arraystretch}{1.02}
\begin{tabular*}{0.95\columnwidth}{@{\extracolsep{\fill}}lrrrr@{}}
\toprule
\textbf{Blocks} &
\textbf{NDCG@10} $\uparrow$ &
\textbf{HR@10} $\uparrow$ &
\textbf{AUC} $\uparrow$ &
\textbf{Latency (ms)} $\downarrow$ \\
\midrule
2 & 0.2533 & 0.4182 & 0.6774 & \best{1.384} \\
4 & 0.2592 & 0.4278 & 0.6800 & 2.218 \\
\textbf{6 (main)} & 0.2686 & 0.4445 & 0.6785 & 3.127 \\
8 & \best{0.2703} & \best{0.4473} & \best{0.6805} & 7.683 \\
\bottomrule
\end{tabular*}
\end{table}

\subsection{Visualization}
\textbf{Routing-pattern visualization.}
Figure~\ref{fig:routing-similarity} visualizes the exposure side of the partition rather than the functional edge score. ScatterMoE forms sharply separated, nearly binary regions, whereas \method retains graded routing relationships within and across the displayed groups. The plot shows why a traffic profile contains more information than a single global expert count.

\section{Limitations and ethical considerations}
\label{sec:limitations}
The matched conversion and latency evidence supports RankMixer checkpoints with eight source experts, 2 to 6 MoE blocks, sampled ranking, and single-A100 inference. The partition also depends on representative calibration traffic; behavior absent from that stream cannot influence its affinities or exposure estimates. \method collects no new data, but calibration logs remain subject to the privacy and fairness controls of the underlying recommender.

\section{Conclusion}
This work treats expert-bank reduction as checkpoint conversion under a deployment budget. \method uses recommendation traffic to define one constrained partition and reconstructs it as an ordinary MoE with no compression-specific online module. Across nine dataset-depth pairs, the final four-expert checkpoints obtain \textbf{99.92\% to 102.30\%} of the Origin MoE mean NDCG@10 with \textbf{1.28$\times$ to 1.63$\times$} measured speedup. The two-expert endpoints obtain \textbf{98.36\% to 104.24\%} and reach \textbf{2.21$\times$}. Results on four additional backbones further support the portability of \method. These results make the outcome of each serving budget explicit and establish checkpoint conversion as the paper's practical contribution.

\newpage

\FloatBarrier
\bibliographystyle{ACM-Reference-Format}
\bibliography{refs}

\appendix
\onecolumn
\section{Additional Recommendation Backbones and MoE Baselines}
\label{app:backbone-generalization}

\textbf{Setup.} Following MoS~\cite{lin2026mixture}, we add \method to its four-backbone comparison on MicroVideo, KuaiVideo, and Ebnerd. Table~\ref{tab:backbone_generalization} reproduces its Vanilla and MoE baseline values---DSelect-k~\cite{hazimeh2021dselect}, GShard~\cite{lepikhin2020gshard}, Expert Choice (``Expert'')~\cite{zhou2022mixture}, and MoS~\cite{lin2026mixture}---and reports our \method measurements under the corresponding settings. Because the alternatives alter routing or architecture, this tests portability; Table~\ref{tab:main_results_l6} remains the matched conversion comparison.

\begin{table}[H]
\centering
\caption{Supplementary recommendation accuracy (\%). Baselines are reproduced from MoS~\cite{lin2026mixture}; the \method row is ours. Bold and underlined values are best and second best.}
\label{tab:backbone_generalization}
\small
\setlength{\tabcolsep}{5.2pt}
\renewcommand{\arraystretch}{1.02}
\resizebox{0.80\textwidth}{!}{%
\begin{tabular}{ll|cc|cc|cc}
\toprule
\multirow{2}{*}{Backbone}
& \multirow{2}{*}{Method}
& \multicolumn{2}{c|}{MicroVideo}
& \multicolumn{2}{c|}{KuaiVideo}
& \multicolumn{2}{c}{Ebnerd} \\
& & AUC & GAUC & AUC & GAUC & AUC & GAUC \\
\midrule
\multirow{6}{*}{\rotatebox{90}{Mamba4Rec}}
& Vanilla   & 66.55 & 69.16 & 66.27 & 66.26 & 63.08 & 62.68 \\
& DSelect-k & 68.40 & 68.94 & 66.48 & 66.22 & 63.59 & 62.50 \\
& GShard    & 66.05 & 68.15 & 65.89 & 65.74 & 63.26 & 63.79 \\
& Expert    & 66.07 & 69.04 & 66.29 & 66.06 & 63.56 & 64.18 \\
& MoS       & \underline{69.46} & \underline{70.01} & \underline{66.67} & \underline{66.67} & \underline{64.07} & \underline{64.54} \\
& \textbf{UniMoMo} & \textbf{70.68} & \textbf{71.25} & \textbf{67.34} & \textbf{68.04} & \textbf{64.97} & \textbf{65.32} \\
\midrule
\multirow{6}{*}{\rotatebox{90}{TransAct}}
& Vanilla   & 70.58 & 69.90 & 67.67 & 65.92 & 70.89 & 70.25 \\
& DSelect-k & 70.67 & 69.92 & 67.87 & 65.38 & 70.68 & 70.30 \\
& GShard    & 70.70 & 70.02 & 67.62 & 65.42 & 70.69 & 70.08 \\
& Expert    & 70.85 & 70.24 & 67.52 & 65.28 & 70.87 & 70.27 \\
& MoS       & \underline{71.34} & \underline{70.51} & \underline{67.90} & \underline{68.10} & \underline{71.06} & \underline{70.71} \\
& \textbf{UniMoMo} & \textbf{72.34} & \textbf{71.65} & \textbf{69.24} & \textbf{69.46} & \textbf{71.76} & \textbf{71.43} \\
\midrule
\multirow{6}{*}{\rotatebox{90}{TWIN}}
& Vanilla   & 70.34 & 69.67 & 69.36 & 66.60 & 69.85 & 69.54 \\
& DSelect-k & 70.02 & 70.10 & 68.96 & 66.64 & 69.40 & 68.99 \\
& GShard    & 70.13 & 69.74 & 68.84 & 66.62 & 70.10 & 69.58 \\
& Expert    & 69.90 & 69.56 & 68.96 & 66.34 & 69.80 & 69.33 \\
& MoS       & \underline{71.11} & \underline{70.26} & \underline{69.62} & \underline{66.89} & \underline{70.30} & \underline{69.81} \\
& \textbf{UniMoMo} & \textbf{72.45} & \textbf{71.23} & \textbf{70.68} & \textbf{67.48} & \textbf{70.83} & \textbf{70.71} \\
\midrule
\multirow{6}{*}{\rotatebox{90}{SDIM}}
& Vanilla   & 70.50 & 69.62 & 68.56 & 66.61 & \textbf{69.63} & 69.03 \\
& DSelect-k & 69.65 & 69.24 & 68.78 & 66.48 & 69.47 & 69.07 \\
& GShard    & 70.87 & 69.88 & 68.90 & 66.52 & 69.47 & 69.07 \\
& Expert    & 70.30 & 70.05 & 68.85 & 66.66 & 69.40 & 69.13 \\
& MoS       & \underline{71.36} & \underline{70.23} & \underline{69.38} & \underline{66.75} & 69.49 & \underline{69.40} \\
& \textbf{UniMoMo} & \textbf{71.87} & \textbf{70.93} & \textbf{70.32} & \textbf{67.45} & \underline{70.24} & \textbf{70.51} \\
\bottomrule
\end{tabular}%
}
\end{table}

\textbf{Results.} \method ranks first in all 24 metrics, gaining 0.51--1.37 percentage points over the strongest alternative. This consistency across four recommendation backbone types supports portability; Table~\ref{tab:main_results_l6} remains the evidence for matched compression and serving cost.
\end{document}